%% file: iclr2027_conference.tex
\documentclass{article}
\usepackage{iclr2027_conference,times}

\input{math_commands.tex}

\usepackage{amsmath,amsfonts,amssymb}
\usepackage{graphicx}
\usepackage{multirow}
\usepackage{amsthm}
\usepackage{mathrsfs}
\usepackage[title]{appendix}
\usepackage{xcolor}
\usepackage{textcomp}
\usepackage{manyfoot}
\usepackage{booktabs}
\usepackage{algorithm}
\usepackage{algorithmicx}
\usepackage{algpseudocode}
\usepackage{listings}
\usepackage{threeparttable}
\usepackage{adjustbox}
\usepackage{hyperref}
\usepackage{url}

\title{Temporal Residual Neural Radiance Fields for Monocular Video Dynamic Human Body Reconstruction}

\author{
Tianle Du, Jie Wang, Xiaolong Xie, Wei Li\thanks{Corresponding author.}, Pengxiang Su, Jie Liu \\
Nanchang University \\
\texttt{8008121161@ncu.edu.cn}
}

\iclrfinalcopy
\begin{document}
\maketitle

\begin{abstract}
In the field of computer vision and graphics, high-quality reconstruction of the human body in static scenes has been achieved in recent years by a single multilayer perceptron (MLP) in a number of approaches. However, MLPs have capacity limitations, requiring substantial training time and computational resources for dynamic scene reconstruction. And the quality of reconstruction is significantly constrained. This paper proposes a method for effectively processing complex spatiotemporal signals in dynamic scene human 3D modeling. The proposed method uses Temporal Residual Neural Radiance Fields to achieve novel view rendering and new pose synthesis of human bodies.To address the problem of representing temporal signals in video sequences, we construct a temporal residual field which is not related to the MLP architecture. Secondly, to improve reconstruction efficiency, we propose an integrated approach that reduces trainable parameters and accelerates rendering, thereby enhancing the network's feature representation capability. Finally, we design a multi-dimensional loss function to accurately measure the loss between predicted and actual spatial pixel values. The experimental results show that our proposed approach improves the peak signal-to-noise ratio (PSNR) and structural similarity index (SSIM) accuracy metrics compared to the latest representative methods. It maintains similar accuracy to Anim-NeRF and Neural Body while achieving a nearly 780-fold increase in time efficiency.
\end{abstract}

\section{Introduction}
\label{sect:intro}  
Neural Radiance Fields (NeRF)\cite{bib4}  have gained significant attention in the fields of computer graphics and computer vision due to their ability to learn the density and color attributes of every spatial point in a scene, resulting in astonishing effects in novel view synthes \cite{bib53} is and 3D reconstruction. The Multi-Layer Perceptron (MLP) neural network structure used by NeRF can represent continuous spatio-temporal signals, demonstrating powerful capabilities in 3D reconstruction. In recent years, NeRF has achieved impressive results in synthesising new views of static scenes. This opens up new possibilities for more realistic virtual reality experiences and scene reconstruction.  

Early methods \cite{bib1}\cite{bib36} relied on multi-view stereo to obtain accurate sequences of 3D meshes. Multi-view stereoscopic methods obtain disparity or depth information by matching and aligning images from multiple viewpoints \cite{bib38}. With the acquired disparity or depth information, a point cloud can be created to represent the three-dimensional shape of the human body. However, methods that compute 3D meshes often struggle to accurately depict complex geometric structures and are less robust when faced with complex poses and movements. Additionally, they may struggle with texture missing or occlusion, resulting in limited photorealism. With the introduction of NeRF, 3D reconstruction based on neural radiance fields has become a popular research focus. The method is capable of generating new and realistic views by learning a continuous 3D representation of the scene, thus overcoming the limitations of traditional methods.

However, NeRF and its variants \cite{bib2}\cite{bib15}\cite{bib18}  require a large number of queries to a deep MLP due to the limited capacity of Multi-Layer Perceptrons (MLP). This time-consuming computation makes them challenging to process large spatio-temporal signals such as long videos or dynamic 3D scenes. Furthermore, these models require training times of several hours due to the dual requirements for differentiable deformation modules and volumetric rendering. This also limits their application to scenarios that demand high rendering efficiency. In a previous study, Christian Reiser et al. \cite{bib2} attempted to expedite rendering by replacing a single large MLP with thousands of smaller MLPs, where each MLP represents only a portion of the scene. However, this approach uses a uniform grid to divide the scene, which may not be optimal for irregularly shaped or detail-varied scenes. To address these challenges, one direct method \cite{bib11}\cite{bib37} to increase network complexity is by augmenting the total number of neurons or layers of the MLP. However, this approach can lead to slower inference and rendering speeds, along with increased GPU memory demands. The reason for this is that the inclusion of parameters leads to an increase in both time and memory requirements \cite{bib37}.

In this paper, we propose an MLP architecture independent building block, TRes-NeRF, to model spatio-temporal fields aimed at solving the fast reconstruction of human performers in monocular videos. Specifically, we incorporate a temporal residual into the neural field and replace the linear layers in the MLP with time-dependent layers. These time-dependent weights are modelled as trainable temporal correlation coefficients and fused with the original layer weights. During the training process, our model is able to learn the time-dependent features in the image sequences, which improves the modelling capability of the spatio-temporal correlation. Compared to traditional MLP architectures \cite{bib13}, our TRes-NeRF has a distinct advantage in spatio-temporal modelling. By adding additional trainable parameters without increasing the network width, it significantly reduces the training time. In addition, TRes-NeRF is able to capture dynamic changes in temporal sequences and integrate them into the reconstruction process. This ability to assimilate temporal information enables our model to more accurately reproduce the actions and postures of human performers.

To further accelerate rendering, our network integrates several key structures. First, for computing pixel colors, we employ a joint module consisting of rigid transformation and a spatial skipping scheme \cite{bib32}. This approach extends the reconstruction of input postures into a dynamic canonical space and swiftly bypasses blank spaces by jumping through grid cells, thereby speeding up rendering. Second, to learn typical shapes and appearances, we utilize an efficient variant of Neural Radiance Fields, Instant-NGP\cite{bib3}, which replaces the Multi-Layer Perceptron with a more efficient hash table as its data structure. And we evaluated our method on both synthetic and real monocular videos of moving human figures. Compared to current state-of-the-art methods, our approach not only demonstrates superior reconstruction quality but also requires significantly shorter training periods.Therefore, our method is more suitable for monocular video human dynamic reconstruction and provides a new idea for human 3D reconstruction.

In summary,our contributions in this study are twofold:

\begin{itemize}
    \item We proposed a residual building block to model the spatio-temporal field that is not related to the MLP architecture, allowing the use of smaller MLPs without sacrificing reconstruction quality, and facilitating accelerated training with reduced GPU memory requirements.

   \item We integrate a method to accelerate the volumetric rendering process that enables fast real-time reconstruction of human performers in monocular videos. And we validate the effectiveness of our method on a number of challenging tasks.
\end{itemize}

\section{Related Work}
\label{sect:Related Work} 
\textbf{Neural Radiance Fields}\quad NeRF\cite{bib4} is a fully connected neural network oriented towards three-dimensional implicit space modeling, designed to reconstruct high-quality 3D scenes from multi-view 2D images, which has become a mainstream method in novel view synthesis and 3D reconstruction. Its core idea is to model the color and density of each 3D point in the scene as a function, without the intermediate process of 3D reconstruction, and only based on the positional parameters with the image, it can be used for new-view image synthesis. However, NeRF requires the evaluation and rendering of a large number of 3D sampling points using a deep MLP, and for each sample point millions of queries to an MLP to obtain the density and brightness. Therefore, rendering temporally and spatially complex scenes efficiently through NeRF presents a challenge. In further research, a key approach has been the use of positional encoding \cite{bib5}\cite{bib6} to increase the effective capacity of the MLP, transforming the modeling of radiance and density into functions of position and viewing direction. However, this approach is not sufficient for accelerated training on large-scale data and complex scenarios due to the increased amount of encoding computation.\\

\noindent
\textbf{Monocular Human Reconstruction}\quad In recent research\cite{bib9}\cite{bib12}, the use of neural representation has become a key technology for reconstructing high-quality 3D human models, especially for generating free-view videos of human performers. Peng et al\cite{bib8} implemented human model animation through neural blending weight fields and skeletal-driven deformation. The method improves the constraints of the optimization algorithm and is able to more normalize the learning of the deformation field to accomplish the reconstruction of a dynamic human body in monocular videos. Noguchi et al\cite{bib10} proposed a deformable 3D representation based on articulated objects, where the algorithm formulates an implicit representation of a 3D articulated object by taking into account the rigidity transformations of the most relevant parts of the object, which in turn renders the changes related to the pose. While these methods can learn human models from monocular videos and achieve realistic reconstructions, the slow speed of canonical representation and deformation algorithms results in relatively slow training and rendering speeds\cite{bib39}\cite{bib40}. In contrast, our method addresses this issue and achieves realistic reconstruction quality.\\

\noindent
\textbf{Temporal Field}\quad In recent years, researchers have attempted various ways to accelerate the inference of traditional NeRF, some of which have achieved real-time rendering performance. Yu et al. \cite{bib14} used sparse 3D grids and spherical harmonics to represent scenes, enhancing the optimization speed in reconstruction by adjusting through gradients and regularization methods from image calibration. However, these methods \cite{bib16}\cite{bib17} are only applicable to static scenes and still pose challenges in modeling dynamic scenes. To address this problem, researchers introduced the concept of temporal fields \cite{bib13}, which can be considered as functions that change over time that describes the motion and deformation of objects in the scene. By introducing the temporal dimension, NeRF can model objects in the scene over time, thus capturing the details and subtle changes in dynamic scenes. The key to using temporal fields in our TRes-NeRF architecture for reconstruction is to treat time as an additional input dimension and input it into the neural network along with spatial coordinates for rendering. In this way, the neural network can learn the correlations between time and space and thus infer temporal scene changes.\\

\noindent
\textbf{Residual Connection}\quad Residual connection \cite{bib21} is a technique to address the vanishing and exploding gradient problems in neural network training \cite{bib22}, proposed and applied to residual neural networks by Kaiming He et al\cite{bib47}. It passes the residuals from the intermediate layers to the subsequent layers by expressing the output of each layer as a linear superposition of the input and the input after a single nonlinear transformation.  This type of connection allows the network to learn the difference between the input and the target and thus train and optimize more efficiently.  It can significantly enhance training performance in the modeling process of NeRF. The recent ReRF \cite{bib23} introduced a method to model the residual information between adjacent timestamps in the modeling space by using a compact motion mesh and a residual feature mesh and exploiting the similarity of inter-frame features, thus effectively modeling dynamic scenes. Our TRes-NeRF layers model the residuals of MLP weights, which is different from simple addition of residuals to MLP layer outputs \cite{bib24}\cite{bib25} and methods that optimize the residuals of model parameters \cite{bib26}\cite{bib27}. This modeling approach enhances the capability of TRes-NeRF's MLP linear layers in neural field representations, making it more suitable for modeling complex real-world spatiotemporal signals.\\

\noindent
\textbf{Accelerating Neural Radiance Fields}\quad With the development of neural representations, many methods based on differentiable rendering \cite{bib4}\cite{bib30} have optimized individual neural representations for each scene. However, this optimization process typically takes several hours on a GPU, resulting in high computational costs. Currently, to optimize training and rendering speeds, some methods like TensoRF\cite{bib28} and NSVF\cite{bib29}focus on replacing the architecture in neural representations with more efficient representations for faster training. Recently, Instant-NGP\cite{bib3} achieved the use of a smaller MLP structure and faster training by using a more efficient hash encoding to replace trigonometric function frequency encoding as the data structure for storing feature grids at different coarse scales. We integrated this multi-resolution hash encoding structure with our TRes-NeRF network to achieve higher quality and faster speed 3D reconstruction of the human body.

\section{Method}
\label{sect:Method} 
\begin{figure}[t!]
    \centering
    \includegraphics[width=1\linewidth]{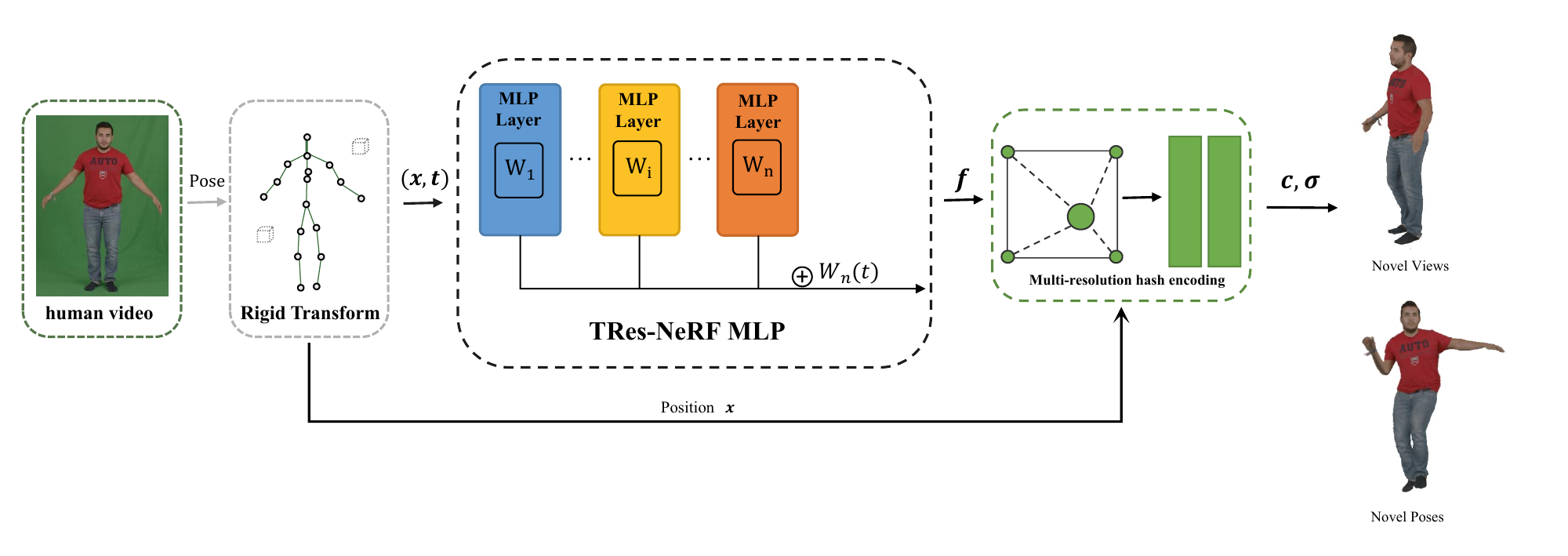}
    \caption{\textbf{Overview of TRes-NeRF Network Structure.}For a given sequence of human videos, after applying rigid transformations to the human joints, we utilize an occupancy grid to filter out points in empty space to reduce the computational load. The remaining points are then extracted by the TRes-NeRF network to extract time-dependent features among the image sequences and are input into a multiresolution hash coding structure along with spatial location features to evaluate their color and density.}
    \label{fig:network structure}
\end{figure}

Given a monocular video of a dynamic human performer, the goal of our model is to achieve fast human model reconstruction with a low-cost computational resource. In this section, we detailed the architecture of our network framework, as illustrated in Fig. \ref{fig:network structure}. We begin by introducing the foundation of our framework (3.1), a linear transformation of the basic temporal field. Subsequently, we described two main components within our framework: a residual neural field for spatiotemporal signals (3.2), which enhances network complexity through the addition of trainable parameters. Following that, we proposed an integrated method for accelerated training (3.3) and completed the reconstruction in a few minutes.

\subsection{Typical Temporal Neural Radiance Field}
\label{sect:Typical Temporal Neural Radiance Field} 
In the temporal neural radiance field, a neural network parameterized by $\Phi _{\theta }$\cite{bib4} is used to encode the scene signal:$\mathbb{R}^{n} \times  \mathbb{R}\longrightarrow  \mathbb{R}^{r}$.Specifically, the neural network takes a spatiotemporal coordinate pair$(x\in \mathbb{R}^{n},t\in \mathbb{R} )$ as input and maps it to a scalar field $f\in \mathbb{R} ^{r}$. The architectural design of the neural network enables it to model viewpoint-related effects. In other words, the definition of the temporal neural radiance field\cite{bib48} can be described as:
\begin{equation}
\Phi _{\theta } (x,t)=\upsilon _{n} (W_{n}(h_{1}\circ h_{2} \circ h_{3}\cdots \circ h_{n-1})(x,t)+b_{n}  )
\end{equation}
\begin{equation}
h_{i} (x_{i} ,t)=\upsilon _{i} (W_{i}x_{i} +b_{i}  )
\end{equation}
In this definition, $h_{i}:\mathbb{R}_{i}^{a}\longrightarrow \mathbb{R}_{i}^{b}$ represents the $i$th layer in the neural network's MLP. The encoding of each layer at spatial position $x_{i} \in \mathbb{R}^{b_{i} }$ is transformed by a weight matrix $W_{i} \in \mathbb{R}^{a_{i}\times b_{i} }$ that has undergone linear transformation, a bias value $b_{i} \in \mathbb{R}^{a_{i} }$, and a nonlinear activation function $\upsilon _{i} $. Through the learning and optimization of the neural network $\Phi _{\theta } $, the temporal neural field is capable of learning the spatiotemporal patterns and dynamic regularities within the input data. This enables the modeling and prediction of continuous signals.

\subsection{Residual Neural Radiance Field}
\label{sect:Residual Neural Radiance Field} 
Traditional MLP structures primarily focus on the fully connected mapping between input features\cite{bib49}, where the number of parameters in each layer grows exponentially with the increase in input dimensions. This leads to a large scale of the network model and constrains its capability in modeling complex spatiotemporal signals. To overcome this challenge, this study introduced a Residual Neural Radiance Field for spatiotemporal signals (Fig.\ref{fig:network structure}), which enhanced the network's ability to model spatiotemporal signals by inserting a special residual connection into the MLP structure. Specifically, we replaced the linear layer in the MLP with a temporal residual layer, which is defined as follows:

\begin{equation}
h_{i} (x_{i} ,t)=\upsilon _{i}((W_{i}+W_{i}(t))x_{i} +b_{i} ) 
\end{equation}

In this structure, $W_{i}(t):\mathbb{R}\to   \mathbb{R}^{a_{i}\times b_{i} } $ introduced are residual parameters related to the time series. This simple substitution enhances the capacity of the MLP by increasing trainable temporal parameters instead of directly increasing the number of neurons, while maintaining the implicit regularization characteristics of the neural network.

\noindent
\textbf{TRes-NeRF Factor Decomposition}\quad After replacing the linear layers of the MLP, if the temporal residual $W_{i}(t)$ optimization is simply added to the MLP as a trainable weights dictionary, the model will introduce many additional parameters in order to capture temporal dependencies,which in turn increases the training and inference computational costs. To address this spatiotemporal signal segmentation issue\cite{bib50}, we decomposed this temporal residual and optimized the $\mathbb{R}_{i} $-dimensional generative set shared by the entire spatiotemporal signal's residual network weights. The optimized definition of the temporal residual is as follows:

\begin{equation}
W_{i}(t)= {\textstyle \sum_{{k=1} }^{R_{i} }}g_{i} (t)[k]\cdot V_{i}[k] 
\end{equation}

In this approach, $g(t)\in \mathbb{R} ^{\mathbb{R}_{i}  } $ and the generative set $V\in \mathbb{R}^{{R}_{i} \times {a}_{i} \times {b}_{i}}$ are identified as trainable parameters related to temporal features. $\left [  \right ] $ are used to denote the selection of elements. By setting $k$ as a matrix, temporal sequence data can be divided into multiple sub-intervals, and linear interpolation is applied to fill the parameter values between these sub-intervals. This method\cite{bib51} not only allows the model to utilize the continuity and dynamic characteristics of data more effectively in the temporal dimension but also significantly reduces the total number of trainable parameters.

\subsection{Additional Accelerated Rendering Methods}
\label{sect:Additional Accelerated Rendering Methods} 
\textbf{Input Stage}\quad Due to the presence of numerous open areas or obstructions around the human body in the input monocular video scenes, computing these areas is a waste of resources. Therefore, to save GPU computational resources, we adopt a rigid transformation and empty voxel skipping scheme \cite{bib32} at the input stage. This method divides the space around the human body into discrete grid cells by maintaining an occupancy grid. For point samples within unoccupied cells, we directly set their density to zero without querying the hypothetical radiance fields, reducing unnecessary computations.The effect of whether or not to use this empty voxel skipping scheme on rendering speed is shown in Table \ref{empty voxel skipping}.

\begin{table}[h]
\begin{center}
\caption{Empty voxel skipping}\label{empty voxel skipping}%
\begin{threeparttable}
\begin{tabular}{@{}lll@{}}
\toprule
& Mean Trainin$\downarrow $  & Mean Rendering$\uparrow$\\
\midrule
w/o empty voxel skipping   & 3m16s  & 7 FPS   \\
w/ empty voxel skipping    & \textbf{1m38s}   & red{\textbf{13 FPS}}   \\
\toprule
\end{tabular}
\begin{tablenotes}[flushleft]
\item We compare training and rendering speeds with and without the use of empty voxel skipping. We report the average training time for 50 epochs in PeopleSnapshot. Training and rendering are significantly faster using the occupied grid jumping scheme.
\end{tablenotes}
\end{threeparttable}
\end{center}
\end{table}

\noindent
\textbf{Rendering Stage}\quad During the training stage, we enhanced the network's ability to model spatiotemporal signals by using TRes-NeRF building blocks. To further achieve rapid rendering and inference, we adopted a recent Instant-NGP\cite{bib3} to parameterize scene signals. By using hash tables to store feature grids at different coarse scales, we eliminate hash collisions and reduce computations. Our method utilizes TRes-NeRF to capture temporal features from image sequences during volumetric rendering. These features are then combined with the original positional features to compute the color and density properties of pixels in space. Such integration enables our model to improve reconstruction quality while achieving faster rendering speeds, thereby optimizing the process of neural rendering.

\subsection{Training Loss}
\label{sect:Training Loss} 
We have adopted a comprehensive loss function framework to train our model, aimed at accurately capturing the three-dimensional structure of the human body in dynamic scenes. This framework takes into account the accuracy of color prediction, transparency handling, and the regularization of the model.

To evaluate the accuracy of color prediction, we calculate the Mean Squared Error (MSE) between the predicted RGB values and the target RGB values, assessing the discrepancy between them:

\begin{equation}
L_{color}=MSE(R,R_{oi} ) 
\end{equation}

Where $R$ symbolizes the predicted RGB values, and $R_{oi}$ represents the corresponding true RGB values of the image.

To reduce floating artifacts in space, we apply a Mean Squared Error (MSE)\cite{bib52} loss to the predicted transparency or visibility:

\begin{equation}
L_{trs}=MSE(\alpha ,\alpha _{oi} ) 
\end{equation}

In our model, to constrain and optimize the processing of neural network outputs for time-involved fine details (tif), thereby enhancing the model's stability and accuracy in handling subtle spatial variations. We apply L2 regularization to the outputs of the tif neural network:

\begin{equation}
L_{L_{2} }=\left \| sigmoid(tif_{op} ) \times\beta  \right \| _{2}^{2} 
\end{equation}

Where $tif_{op}$ represents the output of the neural network, and $\beta$ is a scaling factor used to regulate the impact of the tif neural network output on subsequent operations.

To further enhance the accuracy and efficiency of the model, we employ a combination of multi-dimensional loss functions:

\begin{equation}
L=\omega _{1}L_{color}+  \omega _{2}L_{trs}+\omega _{3}L_{L_{2}}
\end{equation}

With this multi-dimensional loss function combination, our model is capable of efficiently handling dynamic human body modeling tasks in complex three-dimensional scenes, improving rendering quality while maintaining high computational performance.

\section{Experiment}

In order to validate the speed and effectiveness of our method, we conducted experiments on dynamic human body reconstruction from monocular videos and compared them with other state-of-the-art methods, demonstrating the efficacy of our method.

\subsection{Set the Dataset}

\textbf{PeopleSnapshot}\cite{bib30}\quad This dataset is a widely-used benchmark for human monocular video modeling, which includes dynamic videos of humans rotating in front of a camera and provides parameters such as human body masks and keypoints. We evaluated our method on this dataset and conducted a fair comparison with baseline methods.\\

\noindent
\textbf{NeuMan}\cite{bib31}\quad Due to the limitations of the PeopleSnapshot dataset, which provides imperfect posture parameters and only includes simple human rotational movements, there are some constraints in assessing the competitiveness of our method. Therefore, to more accurately evaluate performance in complex postures, we introduced the NeuMan dataset. It offers a richer variety of human postures and movements across different scenes. For each video sequence, we utilized the Openpose algorithm \cite{bib41} for 2D keypoint estimation and employed the Segment-Anything method \cite{bib45} for scene segmentation. Additionally, we used the ROMP algorithm \cite{bib46} to estimate camera parameters and SMPL model parameters, which were then incorporated into the training of our model.

\subsection{Baseline}
To validate our method, we compared it with the following reconstruction techniques:

\noindent
\textbf{Anim-NeRF}\cite{bib7}\quad This baseline uses an MLP-based NeRF for human body reconstruction from videos of individual people. It takes a frame video sequence containing posture changes as input. The SMPL parameters for each frame are first estimated, then sample points along the camera rays in the observation space are rendered. Finally, using posture-guided deformation, these sampling points are transformed back into the canonical space.

\noindent
\textbf{Neural Body}\cite{bib33}\quad This baseline utilizes a statistical human body model to integrate temporal information. By feeding structured latent codes into the network, which are attached to the human body model, the latent code for each point can be obtained by trilinear interpolation of its neighboring points and mapping the density and color values using the MLP network.

\begin{table*}[h]
\setlength{\tabcolsep}{4pt} 
\caption{Qualitative comparison with SoTA methods on the PeopleSnapshot[30] dataset.}\label{peoplesnapshot1}
\begin{center}
\resizebox{\textwidth}{!}{%
\begin{tabular}{@{}lcccccccc@{}}
\toprule%
& \multicolumn{2}{@{}c@{}}{male-3-casual} & \multicolumn{2}{@{}c@{}}{male-4-casual} & \multicolumn{2}{@{}c@{}}{female-3-casual}& \multicolumn{2}{@{}c@{}}{female-4-casual}\\\cmidrule{2-3}\cmidrule{4-5}\cmidrule{6-7}\cmidrule{8-9}%
 & PSNR$\uparrow$ & SSIM$\uparrow$ & PSNR$\uparrow$ &  SSIM$\uparrow$ & PSNR$\uparrow$ &  SSIM$\uparrow$ & PSNR$\uparrow$ & SSIM$\uparrow$\\\midrule
Neural Body\cite{bib33}(14 hours)  & 24.94 & 0.9428 & 24.71 & 0.9496 & 23.87 & 0.9504 & 24.37 & 0.9451\\
\midrule
Anim-NeRF\cite{bib7}(13 hours)   & 29.37 & 0.9703 & \textbf{28.37} & 0.9605 & \textbf{28.91} & \textbf{0.9743} & 28.90 & 0.9678\\
\midrule
InstantAvatar\cite{bib32} (1 minute)  & 29.65 & 0.9730 & 27.97 & 0.9649 & 27.90 & 0.9722 & 28.92 & 0.9692\\
\midrule
Ours\textbf{(1 minute) }  & \textbf{29.89} & \textbf{0.9735} & 28.12 & \textbf{0.9652 }& 28.42 & 0.9728 & \textbf{29.56} & \textbf{0.9714}\\
\toprule%

\end{tabular}
}
{\footnotesize\emph{Note:} The PSNR and SSIM metrics of images generated by TRes-NeRF were compared with those produced by three state-of-the-art methods: Neural Body\cite{bib33}, Anim-NeRF\cite{bib7}, and InstantAvatar\cite{bib32}. This comparison evaluated the speed and accuracy of these four methods. TRes-NeRF achieved a training speed comparable to that of InstantAvatar, while obtaining superior reconstruction quality.\par}
\end{center}
\end{table*}

\textbf{InstantAvatar}\cite{bib32}\quad This baseline takes spatial positions as input and samples along the rays in the located space for each frame. The points are then transformed into normalized space after using the occupancy grid to filter the points in empty voxel space. Finally, the remaining points are deformed into the canonical space using a joint module to evaluate color and density.

\subsection{Compared with SoTA Methods}
\textbf{Reconstruction Quality}\quad For comparison, we animated and rendered the human model for each frame of pose in the PeopleSnapshot dataset. Table \ref{peoplesnapshot1} lists the quantitative analysis of Peak Signal-to-Noise Ratio (PSNR) and Structural Similarity Index (SSIM) across four datasets of the PeopleSnapshot dataset for four of the latest representative methods. The experimental results indicate that, compared to the current state-of-the-art InstantAvatar, our proposed method achieves optimal results in terms of PSNR and SSIM. Additionally, in comparison with Anim-NeRF and Neural Body, our method maintains similar accuracy while enhancing operational efficiency by nearly 780 times. Overall, our proposed method achieves state-of-the-art performance.

\begin{figure}[t!]
    \centering
    \includegraphics[width=1\linewidth]{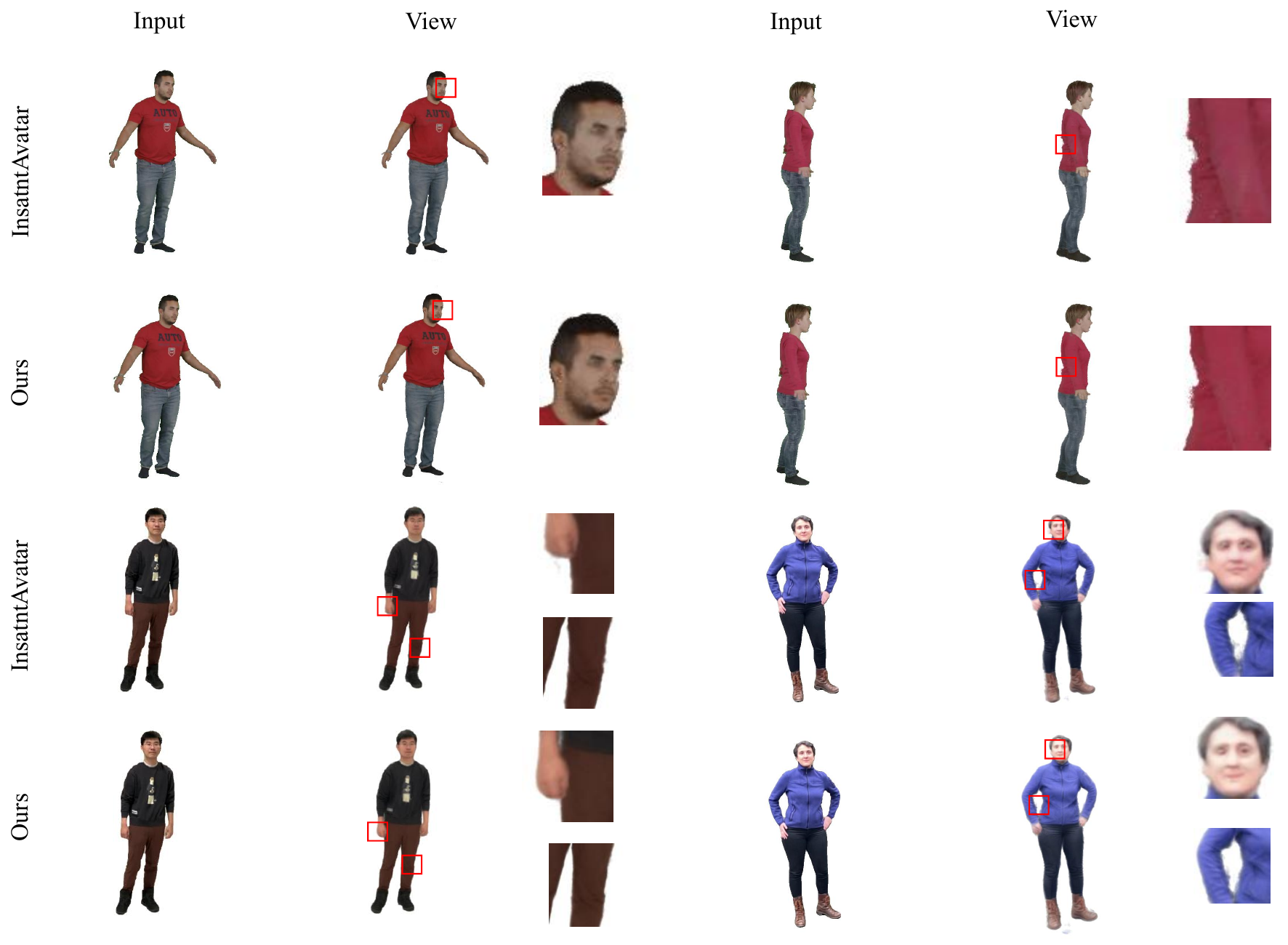}
    \caption{Results of human body reconstruction on the PeopleSnapshot dataset\cite{bib29} (above) and the Neuman dataset\cite{bib32} (below) demonstrate that TRes-NeRF excels in reconstructing details.}
    \label{fig:Detail Comparison}
\end{figure}

\begin{figure}[t!]
    \centering
    \includegraphics[width=1\linewidth]{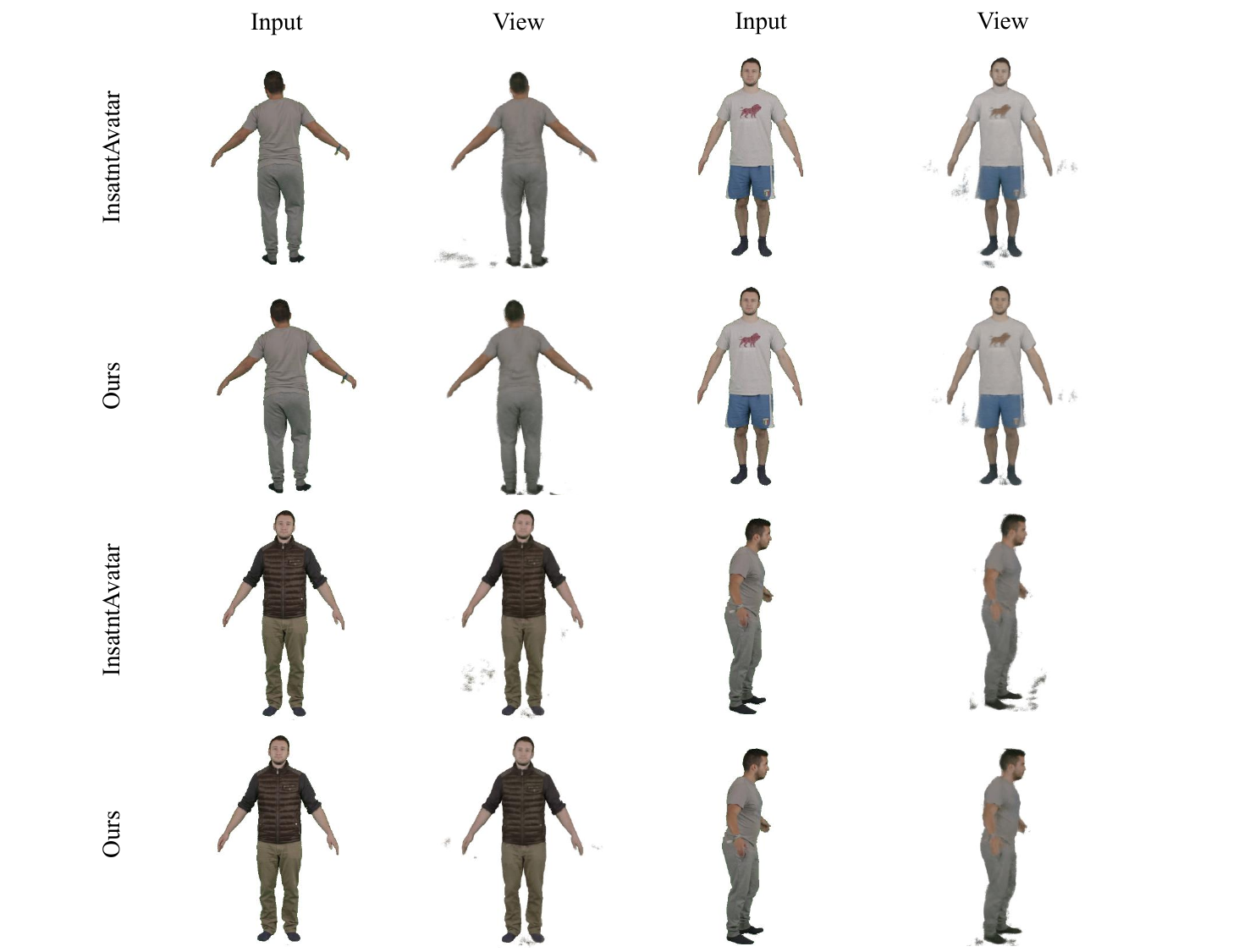}
    \caption{More challenging human body reconstruction results on the PeopleSnapshot\cite{bib30} dataset.The reconstruction results of TRes-NeRF and InstantAvatar on challenging human models in the PeopleSnapshot dataset. The images demonstrate that TRes-NeRF is capable of effectively eliminating artifacts in the reconstruction.}
    \label{fig:artifacts}
\end{figure}

With the aid of TRes-NeRF, we successfully completed the reconstruction of monocular videos at an exceptionally high rendering speed while maintaining the quality of the reconstruction. The results of the image reconstruction quality are shown in Fig. \ref{fig:Detail Comparison}, which indicates that our reconstruction is closer to real images. Our method excels at reconstructing details (as shown in Fig. \ref{fig:Detail Comparison}) and is more effective in eliminating artifacts post-reconstruction (as shown in Fig. \ref{fig:artifacts}). In contrast, InstantAvatar experiences reconstruction failures in detail-rich areas such as the face and legs, and exhibits more artifacts.The TRes-NeRF network can more effectively utilize the continuity and dynamic characteristics of data in the time dimension(sec \ref{sect:Typical Temporal Neural Radiance Field}). And this multi-dimensional loss function framework can improve the stability and accuracy of the model when dealing with subtle spatial changes, making it better at reconstructing details(sec \ref{sect:Training Loss}). Therefore, our method outperforms the baseline.

\begin{table*}[ht]
\caption{Qualitative results on the Neuman dataset.\cite{bib32}}\label{Neuman}
\resizebox{\textwidth}{!}{%
\begin{threeparttable}
\begin{tabular}{@{}lcccccccccccc@{}}
\toprule
& \multicolumn{2}{c}{bike} & \multicolumn{2}{c}{citron} & \multicolumn{2}{c}{jogging}& \multicolumn{2}{c}{lab} & \multicolumn{2}{c}{parkinglot} & \multicolumn{2}{c}{seattle} \\
\cmidrule(lr){2-3}\cmidrule(lr){4-5}\cmidrule(lr){6-7}\cmidrule(lr){8-9}\cmidrule(lr){10-11}\cmidrule(lr){12-13}
 & PSNR$\uparrow$ & SSIM$\uparrow$ & PSNR$\uparrow$ &  SSIM$\uparrow$ & PSNR$\uparrow$ &  SSIM$\uparrow$ & PSNR$\uparrow$ & SSIM$\uparrow$ & PSNR$\uparrow$ & SSIM$\uparrow$ & PSNR$\uparrow$ & SSIM$\uparrow$\\
\midrule
InstantAvatar\cite{bib32}   & 24.18 & \textbf{0.9498} & 24.95 & 0.9491 & 23.40 & 0.9338 & 27.34 & \textbf{0.9731} & 24.14 &0.9520 & 26.57 & \textbf{0.9674}\\
\cmidrule{1-13}
Ours& \textbf{24.32} & 0.9487 & \textbf{24.99} & \textbf{0.9496} & \textbf{23.49} & \textbf{0.9339} & \textbf{27.54} & 0.9697 & \textbf{24.45} &\textbf{0.9528} &\textbf{26.66} & 0.9659\\
\bottomrule
\end{tabular}

\begin{tablenotes}[flushleft]
\item On the Neuman dataset\cite{bib31}, a comparison between images generated by TRes-NeRF and InstantAvatar in terms of PSNR and SSIM was conducted. The data indicates that TRes-NeRF achieved the best PSNR metrics, particularly under more complex motion poses, thereby optimizing the reconstruction quality.
\end{tablenotes}
\end{threeparttable}
}
\end{table*}

\noindent
\textbf{Computational Resources and Speed}\quad Compared to Anim-NeRF \cite{bib7} and Neural Body \cite{bib33}, our method has achieved significant optimization in training time. For instance, while training on an RTX 3090, Anim-NeRF requires 13 hours on 2×RTX 3090 to complete the reconstruction task, and Neural Body needs up to 14 hours on 4×RTX 2080. In contrast, our method requires only one minute on a single RTX 3090, as opposed to several hours, and the reconstruction quality is noticeably superior to these two baseline methods (as indicated in Table \ref{peoplesnapshot1}, Table \ref{Neuman}, and Table \ref{peoplesnapshot2}). Same as InstantAvatar \cite{bib32}, TRes-NeRF adopts a white space hopping scheme as well as multi-resolution hash encoding in the input stage and rendering stage, so the reconstruction index may be closer to stronger. However, TRes-NeRF introduces temporal detail-related information, allowing the network to more accurately capture detailed information in the three-dimensional structure of the human body. This allows our method not only requires less GPU memory and a smaller MLP structure under the same training time budget, but it also achieves noticeably better detailed image quality. These advantages make our method more efficient and practical for human body reconstruction tasks and enable training to be completed at a lower cost.

\begin{table*}[h]
\setlength{\tabcolsep}{0.5pt} 
\caption{Qualitative comparison of more challenging data in the PeopleSnapshot dataset.\cite{bib29}}\label{peoplesnapshot2}
\begin{center}
\resizebox{\textwidth}{!}{%
\begin{tabular}{@{}lcccccccc@{}}
\toprule%
& \multicolumn{2}{@{}c@{}}{male-3-sport} & \multicolumn{2}{@{}c@{}}{male-2-sport} & \multicolumn{2}{@{}c@{}}{male-2-casual}& \multicolumn{2}{@{}c@{}}{female-4-sport}\\\cmidrule{2-3}\cmidrule{4-5}\cmidrule{6-7}\cmidrule{8-9}%
 & PSNR$\uparrow$ & SSIM$\uparrow$ & PSNR$\uparrow$ &  SSIM$\uparrow$ & PSNR$\uparrow$ &  SSIM$\uparrow$ & PSNR$\uparrow$ & SSIM$\uparrow$\\\midrule
InstantAvatar\cite{bib32} (1 minute)  & 23.67 & 0.9216 & \textbf{24.99} & 0.9277 & \textbf{25.25 }& \textbf{0.9306} & 22.04 & 0.8930\\
\midrule
Ours(1 minute)  & \textbf{23.98} & \textbf{0.9281} & 24.92 & \textbf{0.9304} & 25.15 & 0.9270 & \textbf{22.30} & \textbf{0.9021}\\
\toprule%
\end{tabular}
}
{\footnotesize\emph{Note:} A comparison was made between the PSNR and SSIM of images generated by TRes-NeRF and InstantAvatar. The data shows that TRes-NeRF exhibits competitive results in challenging datasets, demonstrating its effectiveness and reliability in processing highly complex and dynamic scenes.\par}
\end{center}
\end{table*}

\begin{figure}[t!]
    \centering
    \includegraphics[width=1\linewidth]{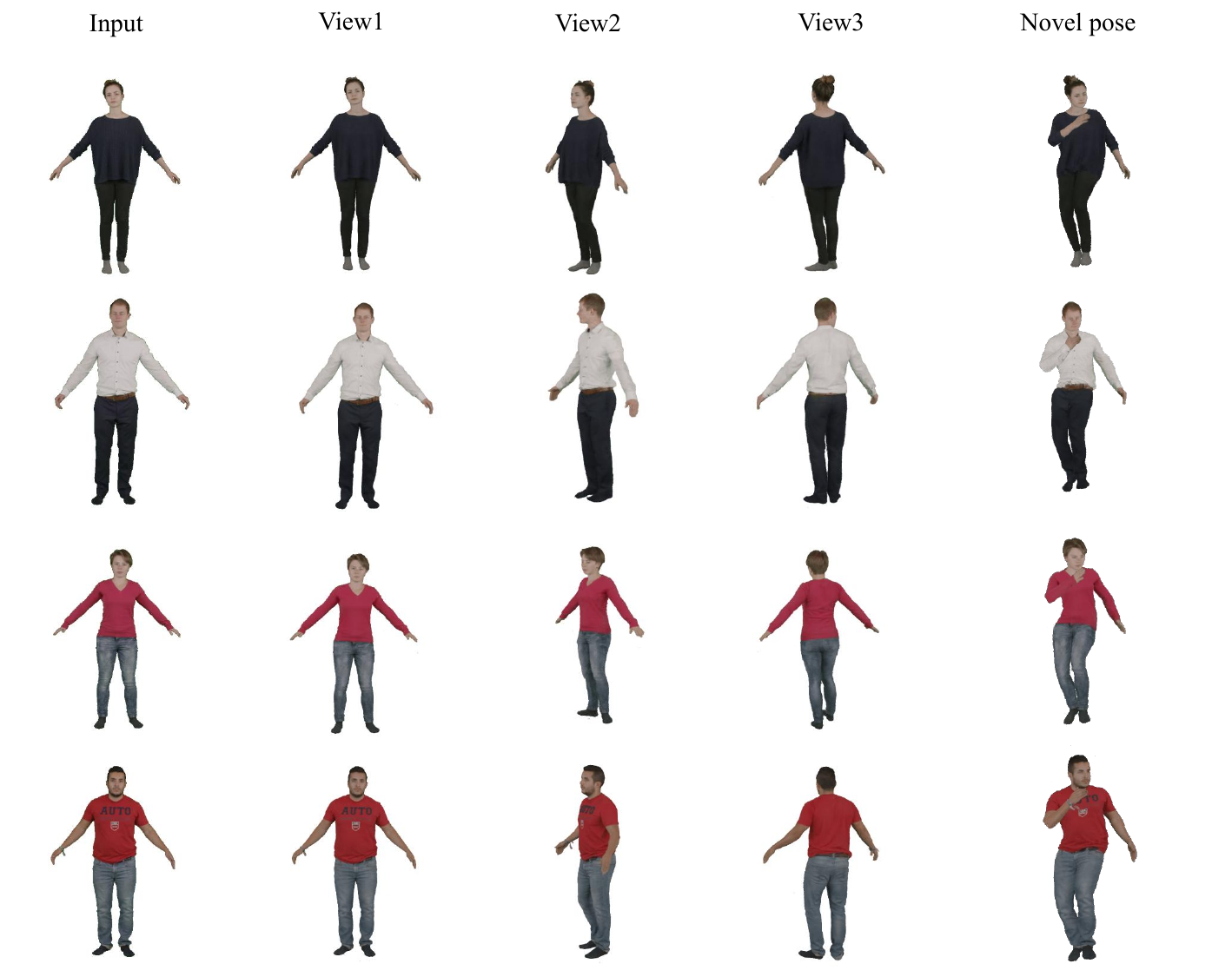}
    \caption{More human body reconstruction results of TRes-NeRF on the PeopleSnapshot dataset\cite{bib29}, including the synthesis effects of new poses.}
    \label{fig:novel pose}
\end{figure}

\noindent
\textbf{Performance in New Pose Synthesis}\quad To verify the effectiveness of the method proposed in this study in the field of new pose synthesis, we conducted challenging experiments on new pose synthesis using our method on the PeopleSnapshot dataset, as shown in Fig. \ref{fig:novel pose}. The results demonstrate that our method performs excellently in handling complex new poses, capable of generating high-quality human animation reconstruction results.

\begin{table}[h]
\begin{center}
\caption{Time residual network}\label{time residual}%
\begin{threeparttable}
\begin{tabular}{@{}llll@{}}
\toprule
& Mean PSNR$\uparrow $  & Mean SSIM$\uparrow$ & GPU$\downarrow $\\
\midrule
w/o Time Residual   & 25.24   & 0.9540  & 3.8G  \\
w/ Time Residual    & \textbf{25.62}   & \textbf{0.9586}  & \textbf{2.4G}  \\
\toprule
\end{tabular}
\begin{tablenotes}[flushleft]
\item A comparison was made on the Neuman dataset between results with and without the temporal residual network. The inclusion of the temporal residual significantly improved reconstruction quality while noticeably reducing GPU memory consumption.
\end{tablenotes}
\end{threeparttable}
\end{center}
\end{table}

\subsection{Ablation Study}
\textbf{Temporal Residual Network}\quad We investigated the impact of the proposed temporal residual layer on the reconstruction effect, as shown in Table \ref{time residual}. After integrating time-related features into the network, it was observed that the network could learn details better and improve the quality of reconstruction. Moreover, in terms of computational resource consumption, the use of temporal residuals effectively saved GPU memory.

\noindent
\textbf{Decomposition of TRes-NeRF}\quad We compared the reconstruction speed when only time-related features were added without employing decomposition techniques, as shown in Table \ref{Decomposition}. The introduction of time-related features brings in more dependencies, and performing decomposition can significantly accelerate both training and rendering speeds.

\begin{table}[h]
\begin{center}
\caption{TRes-NeRF Factor Decomposition}\label{Decomposition}%
\begin{threeparttable}
\begin{tabular}{@{}lll@{}}
\toprule
& Mean Trainin$\downarrow $  & Mean Rendering$\uparrow$\\
\midrule
w/o TRes Factorization   & 4m24s   & 1 FPS   \\
w/ TRes Factorization    & \textbf{1m36s}   & \textbf{13 FPS}   \\
\toprule
\end{tabular}
\begin{tablenotes}[flushleft]
\item In the PeopleSnapshot dataset, a comparison was made of the training and rendering speeds with and without factor decomposition. After decomposing time-related features, both training and rendering speeds were significantly accelerated.
\end{tablenotes}
\end{threeparttable}
\end{center}
\end{table}

\section{Limitations}
Although our method can quickly reconstruct high-quality human models from monocular video at low hardware cost, it still has some limitations. Our approach can be helpful when the challenge is cost and network capacity. However, when it comes to addressing the lack of unsupervised constraints, our method may not show advantages in challenging scenarios.

\section{Conclusion}

In this paper, we propose a method for modelling complex spatio-temporal signals and accelerating rendering, with application to fast reconstruction of human monocular videos. Our key idea is to introduce temporal residuals into the neural field, which enhances the ability of the neural network to model spatiotemporal signals by inserting a special residual connection within the MLP structure. This allows us to increase network complexity without adding more neurons, while improving reconstruction quality with smaller MLPs and reduced GPU memory usage. In addition, to achieve faster rendering speeds, we applied multi-resolution hash coding \cite{bib3} during the rendering phase, which simultaneously maps temporal and spatial features to improve our rendering speed. Compared to baseline methods, our approach achieves faster inference and training times with lower GPU memory requirements, resulting in higher quality reconstructions.

\subsection* {Acknowledgement} 
This work was supported by the National Natural Sci-ence Foundation of China (Grant No.62361041; 62262040.


\bibliography{report}   
\bibliographystyle{iclr2027_conference}

\end{document}

%% file: math_commands.tex
\usepackage{amsmath,amsfonts,bm}

\def\eqref#1{equation~\ref{#1}}

\def\1{\bm{1}}

\DeclareMathAlphabet{\mathsfit}{\encodingdefault}{\sfdefault}{m}{sl}
\SetMathAlphabet{\mathsfit}{bold}{\encodingdefault}{\sfdefault}{bx}{n}













%% file: iclr2027_conference.bbl
\begin{thebibliography}{46}
\providecommand{\natexlab}[1]{#1}
\providecommand{\url}[1]{\texttt{#1}}
\expandafter\ifx\csname urlstyle\endcsname\relax
  \providecommand{\doi}[1]{doi: #1}\else
  \providecommand{\doi}{doi: \begingroup \urlstyle{rm}\Url}\fi

\bibitem[Alldieck et~al.(2018)Alldieck, Magnor, Xu, Theobalt, and Pons-Moll]{bib30}
Thiemo Alldieck, Marcus Magnor, Weipeng Xu, Christian Theobalt, and Gerard Pons-Moll.
\newblock Detailed human avatars from monocular video.
\newblock In \emph{2018 International Conference on 3D Vision (3DV)}, pp.\  98--109. IEEE, 2018.

\bibitem[Cao et~al.(2017)Cao, Simon, Wei, and Sheikh]{bib41}
Zhe Cao, Tomas Simon, Shih-En Wei, and Yaser Sheikh.
\newblock Realtime multi-person 2d pose estimation using part affinity fields.
\newblock In \emph{Proceedings of the IEEE conference on computer vision and pattern recognition}, pp.\  7291--7299, 2017.

\bibitem[Chen et~al.(2022)Chen, Xu, Geiger, Yu, and Su]{bib28}
Anpei Chen, Zexiang Xu, Andreas Geiger, Jingyi Yu, and Hao Su.
\newblock Tensorf: Tensorial radiance fields.
\newblock In \emph{European Conference on Computer Vision}, pp.\  333--350. Springer, 2022.

\bibitem[Chen et~al.(2021)Chen, Zhang, Kang, Zhe, Bao, Jia, and Lu]{bib7}
Jianchuan Chen, Ying Zhang, Di~Kang, Xuefei Zhe, Linchao Bao, Xu~Jia, and Huchuan Lu.
\newblock Animatable neural radiance fields from monocular rgb videos.
\newblock \emph{arXiv preprint arXiv:2106.13629}, 2021.

\bibitem[Choi et~al.(2022)Choi, Moon, Armando, Leroy, Lee, and Rogez]{bib40}
Hongsuk Choi, Gyeongsik Moon, Matthieu Armando, Vincent Leroy, Kyoung~Mu Lee, and Gr{\'e}gory Rogez.
\newblock Mononhr: Monocular neural human renderer.
\newblock In \emph{2022 International Conference on 3D Vision (3DV)}, pp.\  242--251. IEEE, 2022.

\bibitem[Collet et~al.(2015)Collet, Chuang, Sweeney, Gillett, Evseev, Calabrese, Hoppe, Kirk, and Sullivan]{bib1}
Alvaro Collet, Ming Chuang, Pat Sweeney, Don Gillett, Dennis Evseev, David Calabrese, Hugues Hoppe, Adam Kirk, and Steve Sullivan.
\newblock High-quality streamable free-viewpoint video.
\newblock \emph{ACM Transactions on Graphics (ToG)}, 34\penalty0 (4):\penalty0 1--13, 2015.

\bibitem[Cook(1984)]{bib36}
Robert~L Cook.
\newblock Shade trees.
\newblock In \emph{Proceedings of the 11th annual conference on Computer graphics and interactive techniques}, pp.\  223--231, 1984.

\bibitem[Dettmers et~al.(2023)Dettmers, Pagnoni, Holtzman, and Zettlemoyer]{bib27}
Tim Dettmers, Artidoro Pagnoni, Ari Holtzman, and Luke Zettlemoyer.
\newblock Qlora: Efficient finetuning of quantized llms.
\newblock \emph{arXiv preprint arXiv:2305.14314}, 2023.

\bibitem[Fridovich-Keil et~al.(2022)Fridovich-Keil, Yu, Tancik, Chen, Recht, and Kanazawa]{bib14}
Sara Fridovich-Keil, Alex Yu, Matthew Tancik, Qinhong Chen, Benjamin Recht, and Angjoo Kanazawa.
\newblock Plenoxels: Radiance fields without neural networks.
\newblock In \emph{Proceedings of the IEEE/CVF Conference on Computer Vision and Pattern Recognition}, pp.\  5501--5510, 2022.

\bibitem[Gao et~al.(2022)Gao, Yang, Kim, Peng, Liu, and Tong]{bib39}
Xiangjun Gao, Jiaolong Yang, Jongyoo Kim, Sida Peng, Zicheng Liu, and Xin Tong.
\newblock Mps-nerf: Generalizable 3d human rendering from multiview images.
\newblock \emph{IEEE Transactions on Pattern Analysis and Machine Intelligence}, pp.\  1--12, 2022.

\bibitem[Garbin et~al.(2021)Garbin, Kowalski, Johnson, Shotton, and Valentin]{bib15}
Stephan~J Garbin, Marek Kowalski, Matthew Johnson, Jamie Shotton, and Julien Valentin.
\newblock Fastnerf: High-fidelity neural rendering at 200fps.
\newblock In \emph{Proceedings of the IEEE/CVF International Conference on Computer Vision}, pp.\  14346--14355, 2021.

\bibitem[Gunst \& Mason(1977)Gunst and Mason]{bib52}
Richard~F Gunst and Robert~L Mason.
\newblock Biased estimation in regression: an evaluation using mean squared error.
\newblock \emph{Journal of the American Statistical Association}, pp.\  616--628, 1977.

\bibitem[He et~al.(2016{\natexlab{a}})He, Zhang, Ren, and Sun]{bib25}
Kaiming He, Xiangyu Zhang, Shaoqing Ren, and Jian Sun.
\newblock Identity mappings in deep residual networks.
\newblock In \emph{Computer Vision--ECCV 2016: 14th European Conference, Amsterdam, The Netherlands, October 11--14, 2016, Proceedings, Part IV 14}, pp.\  630--645. Springer, 2016{\natexlab{a}}.

\bibitem[He et~al.(2016{\natexlab{b}})He, Zhang, Ren, and Sun]{bib47}
Kaiming He, Xiangyu Zhang, Shaoqing Ren, and Jian Sun.
\newblock Deep residual learning for image recognition.
\newblock In \emph{Proceedings of the IEEE conference on computer vision and pattern recognition}, pp.\  770--778, 2016{\natexlab{b}}.

\bibitem[Hedman et~al.(2021)Hedman, Srinivasan, Mildenhall, Barron, and Debevec]{bib16}
Peter Hedman, Pratul~P Srinivasan, Ben Mildenhall, Jonathan~T Barron, and Paul Debevec.
\newblock Baking neural radiance fields for real-time view synthesis.
\newblock In \emph{Proceedings of the IEEE/CVF International Conference on Computer Vision}, pp.\  5875--5884, 2021.

\bibitem[Hochreiter(1998)]{bib22}
Sepp Hochreiter.
\newblock The vanishing gradient problem during learning recurrent neural nets and problem solutions.
\newblock \emph{International Journal of Uncertainty, Fuzziness and Knowledge-Based Systems}, 6\penalty0 (02):\penalty0 107--116, 1998.

\bibitem[Jiang et~al.(2022{\natexlab{a}})Jiang, Hong, Bao, and Zhang]{bib9}
Boyi Jiang, Yang Hong, Hujun Bao, and Juyong Zhang.
\newblock Selfrecon: Self reconstruction your digital avatar from monocular video.
\newblock In \emph{Proceedings of the IEEE/CVF Conference on Computer Vision and Pattern Recognition}, pp.\  5605--5615, 2022{\natexlab{a}}.

\bibitem[Jiang et~al.(2023)Jiang, Chen, Song, and Hilliges]{bib32}
Tianjian Jiang, Xu~Chen, Jie Song, and Otmar Hilliges.
\newblock Instantavatar: Learning avatars from monocular video in 60 seconds.
\newblock In \emph{Proceedings of the IEEE/CVF Conference on Computer Vision and Pattern Recognition}, pp.\  16922--16932, 2023.

\bibitem[Jiang et~al.(2022{\natexlab{b}})Jiang, Yi, Samei, Tuzel, and Ranjan]{bib31}
Wei Jiang, Kwang~Moo Yi, Golnoosh Samei, Oncel Tuzel, and Anurag Ranjan.
\newblock Neuman: Neural human radiance field from a single video.
\newblock In \emph{European Conference on Computer Vision}, pp.\  402--418. Springer, 2022{\natexlab{b}}.

\bibitem[Kalantari et~al.(2016)Kalantari, Wang, and Ramamoorthi]{bib53}
Nima~Khademi Kalantari, Ting-Chun Wang, and Ravi Ramamoorthi.
\newblock Learning-based view synthesis for light field cameras.
\newblock \emph{ACM Transactions on Graphics (TOG)}, 35\penalty0 (6):\penalty0 1--10, 2016.

\bibitem[Karimi~Mahabadi et~al.(2021)Karimi~Mahabadi, Henderson, and Ruder]{bib26}
Rabeeh Karimi~Mahabadi, James Henderson, and Sebastian Ruder.
\newblock Compacter: Efficient low-rank hypercomplex adapter layers.
\newblock \emph{Advances in Neural Information Processing Systems}, 34:\penalty0 1022--1035, 2021.

\bibitem[Kirillov et~al.(2023)Kirillov, Mintun, Ravi, Mao, Rolland, Gustafson, Xiao, Whitehead, Berg, Lo, et~al.]{bib45}
Alexander Kirillov, Eric Mintun, Nikhila Ravi, Hanzi Mao, Chloe Rolland, Laura Gustafson, Tete Xiao, Spencer Whitehead, Alexander~C Berg, Wan-Yen Lo, et~al.
\newblock Segment anything.
\newblock \emph{arXiv preprint arXiv:2304.02643}, 2023.

\bibitem[Liu et~al.(2020)Liu, Gu, Zaw~Lin, Chua, and Theobalt]{bib29}
Lingjie Liu, Jiatao Gu, Kyaw Zaw~Lin, Tat-Seng Chua, and Christian Theobalt.
\newblock Neural sparse voxel fields.
\newblock \emph{Advances in Neural Information Processing Systems}, 33:\penalty0 15651--15663, 2020.

\bibitem[Mildenhall et~al.(2021)Mildenhall, Srinivasan, Tancik, Barron, Ramamoorthi, and Ng]{bib4}
Ben Mildenhall, Pratul~P Srinivasan, Matthew Tancik, Jonathan~T Barron, Ravi Ramamoorthi, and Ren Ng.
\newblock Nerf: Representing scenes as neural radiance fields for view synthesis.
\newblock \emph{Communications of the ACM}, 65\penalty0 (1):\penalty0 99--106, 2021.

\bibitem[M{\"u}ller et~al.(2022)M{\"u}ller, Evans, Schied, and Keller]{bib3}
Thomas M{\"u}ller, Alex Evans, Christoph Schied, and Alexander Keller.
\newblock Instant neural graphics primitives with a multiresolution hash encoding.
\newblock \emph{ACM Transactions on Graphics (ToG)}, 41\penalty0 (4):\penalty0 1--15, 2022.

\bibitem[Niemeyer \& Geiger(2021)Niemeyer and Geiger]{bib17}
Michael Niemeyer and Andreas Geiger.
\newblock Giraffe: Representing scenes as compositional generative neural feature fields.
\newblock In \emph{Proceedings of the IEEE/CVF Conference on Computer Vision and Pattern Recognition}, pp.\  11453--11464, 2021.

\bibitem[Noguchi et~al.(2021)Noguchi, Sun, Lin, and Harada]{bib10}
Atsuhiro Noguchi, Xiao Sun, Stephen Lin, and Tatsuya Harada.
\newblock Neural articulated radiance field.
\newblock In \emph{Proceedings of the IEEE/CVF International Conference on Computer Vision}, pp.\  5762--5772, 2021.

\bibitem[Orbach(1962)]{bib21}
J~Orbach.
\newblock Principles of neurodynamics. perceptrons and the theory of brain mechanisms.
\newblock \emph{Archives of General Psychiatry}, 7\penalty0 (3):\penalty0 218--219, 1962.

\bibitem[Pan et~al.(2017)Pan, Hu, and Cao]{bib51}
Zhuokun Pan, Yueming Hu, and Bin Cao.
\newblock Construction of smooth daily remote sensing time series data: A higher spatiotemporal resolution perspective.
\newblock \emph{Open Geospatial Data, Software and Standards}, 2\penalty0 (1):\penalty0 1--11, 2017.

\bibitem[Peng et~al.(2021{\natexlab{a}})Peng, Dong, Wang, Zhang, Shuai, Bao, and Zhou]{bib8}
Sida Peng, Junting Dong, Qianqian Wang, Shangzhan Zhang, Qing Shuai, Hujun Bao, and Xiaowei Zhou.
\newblock Animatable neural radiance fields for human body modeling.
\newblock \emph{arXiv preprint arXiv:2105.02872}, 2\penalty0 (3):\penalty0 5, 2021{\natexlab{a}}.

\bibitem[Peng et~al.(2021{\natexlab{b}})Peng, Zhang, Xu, Wang, Shuai, Bao, and Zhou]{bib33}
Sida Peng, Yuanqing Zhang, Yinghao Xu, Qianqian Wang, Qing Shuai, Hujun Bao, and Xiaowei Zhou.
\newblock Neural body: Implicit neural representations with structured latent codes for novel view synthesis of dynamic humans.
\newblock In \emph{Proceedings of the IEEE/CVF Conference on Computer Vision and Pattern Recognition}, pp.\  9054--9063, 2021{\natexlab{b}}.

\bibitem[Qi et~al.(2016)Qi, Su, Nie{\ss}ner, Dai, Yan, and Guibas]{bib38}
Charles~R Qi, Hao Su, Matthias Nie{\ss}ner, Angela Dai, Mengyuan Yan, and Leonidas~J Guibas.
\newblock Volumetric and multi-view cnns for object classification on 3d data.
\newblock In \emph{Proceedings of the IEEE conference on computer vision and pattern recognition}, pp.\  5648--5656, 2016.

\bibitem[Raj et~al.(2022)Raj, Verma, and Nagarajan]{bib48}
Ashish Raj, Parul Verma, and Srikantan Nagarajan.
\newblock Structure-function models of temporal, spatial, and spectral characteristics of non-invasive whole brain functional imaging.
\newblock \emph{Frontiers in neuroscience}, 16:\penalty0 959557, 2022.

\bibitem[Reiser et~al.(2021)Reiser, Peng, Liao, and Geiger]{bib2}
Christian Reiser, Songyou Peng, Yiyi Liao, and Andreas Geiger.
\newblock Kilonerf: Speeding up neural radiance fields with thousands of tiny mlps.
\newblock In \emph{Proceedings of the IEEE/CVF International Conference on Computer Vision}, pp.\  14335--14345, 2021.

\bibitem[Shao et~al.(2023)Shao, Zheng, Tu, Liu, Zhang, and Liu]{bib50}
Ruizhi Shao, Zerong Zheng, Hanzhang Tu, Boning Liu, Hongwen Zhang, and Yebin Liu.
\newblock Tensor4d: Efficient neural 4d decomposition for high-fidelity dynamic reconstruction and rendering.
\newblock In \emph{Proceedings of the IEEE/CVF Conference on Computer Vision and Pattern Recognition}, pp.\  16632--16642, 2023.

\bibitem[Sitzmann et~al.(2020)Sitzmann, Martel, Bergman, Lindell, and Wetzstein]{bib13}
Vincent Sitzmann, Julien Martel, Alexander Bergman, David Lindell, and Gordon Wetzstein.
\newblock Implicit neural representations with periodic activation functions.
\newblock \emph{Advances in neural information processing systems}, 33:\penalty0 7462--7473, 2020.

\bibitem[Srivastava et~al.(2015)Srivastava, Greff, and Schmidhuber]{bib24}
Rupesh~Kumar Srivastava, Klaus Greff, and J{\"u}rgen Schmidhuber.
\newblock Highway networks.
\newblock \emph{arXiv preprint arXiv:1505.00387}, 2015.

\bibitem[Su et~al.(2021)Su, Yu, Zollh{\"o}fer, and Rhodin]{bib37}
Shih-Yang Su, Frank Yu, Michael Zollh{\"o}fer, and Helge Rhodin.
\newblock A-nerf: Articulated neural radiance fields for learning human shape, appearance, and pose.
\newblock \emph{Advances in Neural Information Processing Systems}, 34:\penalty0 12278--12291, 2021.

\bibitem[Sun et~al.(2023)Sun, Bao, Liu, Mei, and Black]{bib46}
Yu~Sun, Qian Bao, Wu~Liu, Tao Mei, and Michael~J Black.
\newblock Trace: 5d temporal regression of avatars with dynamic cameras in 3d environments.
\newblock In \emph{Proceedings of the IEEE/CVF Conference on Computer Vision and Pattern Recognition}, pp.\  8856--8866, 2023.

\bibitem[Tancik et~al.(2020)Tancik, Srinivasan, Mildenhall, Fridovich-Keil, Raghavan, Singhal, Ramamoorthi, Barron, and Ng]{bib5}
Matthew Tancik, Pratul Srinivasan, Ben Mildenhall, Sara Fridovich-Keil, Nithin Raghavan, Utkarsh Singhal, Ravi Ramamoorthi, Jonathan Barron, and Ren Ng.
\newblock Fourier features let networks learn high frequency functions in low dimensional domains.
\newblock \emph{Advances in Neural Information Processing Systems}, 33:\penalty0 7537--7547, 2020.

\bibitem[Tolstikhin et~al.(2021)Tolstikhin, Houlsby, Kolesnikov, Beyer, Zhai, Unterthiner, Yung, Steiner, Keysers, Uszkoreit, et~al.]{bib49}
Ilya~O Tolstikhin, Neil Houlsby, Alexander Kolesnikov, Lucas Beyer, Xiaohua Zhai, Thomas Unterthiner, Jessica Yung, Andreas Steiner, Daniel Keysers, Jakob Uszkoreit, et~al.
\newblock Mlp-mixer: An all-mlp architecture for vision.
\newblock \emph{Advances in neural information processing systems}, 34:\penalty0 24261--24272, 2021.

\bibitem[Vaswani et~al.(2017)Vaswani, Shazeer, Parmar, Uszkoreit, Jones, Gomez, Kaiser, and Polosukhin]{bib6}
Ashish Vaswani, Noam Shazeer, Niki Parmar, Jakob Uszkoreit, Llion Jones, Aidan~N Gomez, {\L}ukasz Kaiser, and Illia Polosukhin.
\newblock Attention is all you need.
\newblock \emph{Advances in neural information processing systems}, 30, 2017.

\bibitem[Wang et~al.(2023)Wang, Hu, He, Wang, Yu, Tuytelaars, Xu, and Wu]{bib23}
Liao Wang, Qiang Hu, Qihan He, Ziyu Wang, Jingyi Yu, Tinne Tuytelaars, Lan Xu, and Minye Wu.
\newblock Neural residual radiance fields for streamably free-viewpoint videos.
\newblock In \emph{Proceedings of the IEEE/CVF Conference on Computer Vision and Pattern Recognition}, pp.\  76--87, 2023.

\bibitem[Weng et~al.(2022)Weng, Curless, Srinivasan, Barron, and Kemelmacher-Shlizerman]{bib11}
Chung-Yi Weng, Brian Curless, Pratul~P Srinivasan, Jonathan~T Barron, and Ira Kemelmacher-Shlizerman.
\newblock Humannerf: Free-viewpoint rendering of moving people from monocular video.
\newblock In \emph{Proceedings of the IEEE/CVF conference on computer vision and pattern Recognition}, pp.\  16210--16220, 2022.

\bibitem[Yu et~al.(2021)Yu, Li, Tancik, Li, Ng, and Kanazawa]{bib18}
Alex Yu, Ruilong Li, Matthew Tancik, Hao Li, Ren Ng, and Angjoo Kanazawa.
\newblock Plenoctrees for real-time rendering of neural radiance fields.
\newblock In \emph{Proceedings of the IEEE/CVF International Conference on Computer Vision}, pp.\  5752--5761, 2021.

\bibitem[Zheng et~al.(2021)Zheng, Yu, Liu, and Dai]{bib12}
Zerong Zheng, Tao Yu, Yebin Liu, and Qionghai Dai.
\newblock Pamir: Parametric model-conditioned implicit representation for image-based human reconstruction.
\newblock \emph{IEEE transactions on pattern analysis and machine intelligence}, 44\penalty0 (6):\penalty0 3170--3184, 2021.

\end{thebibliography}
